\documentclass[times,twocolumn,final,nopreprintline]{elsarticle}

\usepackage[utf8]{inputenc}
\usepackage{amsmath,amssymb,amsthm}
\usepackage{framed,multirow}
\usepackage{adjustbox}
\usepackage{dsfont}
\usepackage{hyperref}
\usepackage{color}
\usepackage{makecell}
\usepackage{bbm}
\usepackage{bm}
\usepackage{url}
\usepackage{xcolor}
\usepackage{subcaption}
\usepackage{comment}
\begin{document}

% \verso{Meng Xia \textit{et~al.}}

\begin{frontmatter}

\title{Malignancy Prediction and Lesion Identification from Clinical Dermatological Images}
%\tnoteref{tnote1}}%
%\tnotetext[tnote1]{This is an example for title footnote coding.}

\author[1]{Meng Xia\corref{cor1}}
\cortext[cor1]{Corresponding author}
%\ead{meng.xia@duke.edu}
\author[2]{Meenal K. Kheterpal}
\author[3]{Samantha C. Wong}
\author[3]{Christine Park}
\author[4]{William Ratliff}
\author[1]{Lawrence Carin}
\author[1]{Ricardo Henao}

\address[1]{Department of Electrical and Computer Engineering, Duke University, Durham, USA}
\address[2]{Department of Dermatology, Duke University, Durham, USA}
\address[3]{Duke University, School of Medicine, Durham, NC, USA}
\address[4]{Duke Institute for Health Innovation, Duke University, Durham, NC, USA}

%\received{1 May 2013}
%\finalform{10 May 2013}
%\accepted{13 May 2013}
%\availableonline{15 May 2013}
%\communicated{S. Sarkar}

\begin{abstract}
We consider machine-learning-based malignancy prediction and lesion identification from clinical dermatological images, which can be indistinctly acquired via smartphone or dermoscopy capture.
Additionally, we do not assume that images contain single lesions, thus the framework supports both focal or wide-field images.
Specifically, we propose a two-stage approach in which we first identify all lesions present in the image regardless of sub-type or likelihood of malignancy, then it estimates their likelihood of malignancy, and through aggregation, it also generates an image-level likelihood of malignancy that can be used for high-level screening processes.
Further, we consider augmenting the proposed approach with clinical covariates (from electronic health records) and publicly available data (the ISIC dataset).
Comprehensive experiments validated on an independent test dataset demonstrate that $i$) the proposed approach outperforms alternative model architectures; $ii$) the model based on images outperforms a pure clinical model by a large margin, and the combination of images and clinical data does not significantly improves over the image-only model; and $iii$) the proposed framework offers comparable performance in terms of malignancy classification relative to three board certified dermatologists with different levels of experience.
\end{abstract}


\end{frontmatter}

%\linenumbers
%\maketitle

\section{Introduction}
Prior to the COVID-19 pandemic, access to dermatology care was challenging due to limited supply and increasing demand.
According to a survey study of dermatologists, the mean $\pm$ standard deviation (SD) waiting time was 33$\pm$32 days, 64\% of the appointments exceeded the criterion cutoff of 3 weeks and 63\% of the appointments exceeded 2-week criterion cutoff for established patients.
During the COVID-19 pandemic, the number of dermatology consultations were reduced by 80-90\% to urgent issues only, leading to delay in care of dermatologic concerns.
Moreover, the issue of access is very significant for the growing Medicare population, expected to account for 1 in 5 patients by 2030 \cite{vincent2010next}, due to a higher incidence of skin cancer.

Access issues in dermatology are concerning as there has been an increasing incidence of skin cancers, particularly a 3-fold increase in melanoma over the last 40 years \cite{statfacts}.
Many of the skin lesions of concern are screened by primary care physicians (PCPs).
In fact, up to one third of primary care visits contend with at least one skin problem, and skin tumors are the most common reason for referral to dermatology \cite{lowell2001dermatology}.
High volume of referrals places a strain on specialty care, delaying visits for high-risk cases. 
Given the expected rise in baby boomers, with significantly increased risk of skin cancer, there is an urgent need to equip primary care providers to help screen and risk stratify patients in real time, high quality and cost-conscious fashion.
PCPs have variable experience and training in dermatology, causing often low concordance between their evaluation and dermatology \cite{lowell2001dermatology}.
A consistent clinical decision support (CDS) system has the potential to mitigate this variability, and to create a powerful risk stratification tool, leveraging the frontline network of providers to enhance access to quality and valuable care.
In addition, such a tool can aid tele-dermatology workflows that have emerged during the global pandemic.

Over the last decade, several studies in the field of dermatology have demonstrated the promise of deep learning models such as convolutional neural networks (CNN) in terms of classification of skin lesions \cite{esteva2017dermatologist,haenssle2018man}, with dermoscopy-based machine learning (ML) algorithms reaching sensitivities and specificities for melanoma diagnosis at 87.6\% (95\% CI 72.72-100.0) and 83.5\% (95\% CI: 60.92-100.0), respectively, by meta-analysis \cite{safran2018machine}.
Several authors have reported superior performance of ML algorithms for classification of squamous cell carcinoma (SCC) and basal cell carcinomas (BCC) with larger datasets improving performance \cite{han2018classification,esteva2017dermatologist}. 

From a machine-learning methods perspective, a common approach for classification with dermoscopy images consists on refining pre-trained CNN architectures such as VGG16 as in \cite{lopez2017skin} or AlexNet after image pre-processing, \emph{e.g.} background removal, as in \cite{salido2018using}.
Alternatively, some approaches consider lesion sub-types independently \cite{polat2020detection}, sonified images \cite{dascalu2019skin}, or by combining clinical data with images to increase the information available to the model for prediction \cite{tognetti2021new}.
However, dermoscopy images are generally of good quality, high resolution and minimal background noise, making them less challenging to recognize compared to clinical, wide-field, images.

Beyond dermoscopy images, similar refinement approaches have been proposed based on architectures such as ResNet152 \cite{han2018classification,fujisawa2019deep}, with additional pre-processing (illumination correction) \cite{nasr2016melanoma}, by using detection models to account for the non-informative background \cite{jafari2016skin,jinnai2020development}, or by first extracting features with CNN-based models, \emph{e.g.}, Inception v2, to then perform feature classification with other machine learning methods \cite{dascalu2019skin}.
Moreover, comparative studies \cite{brinker2019convolutional,haenssle2018man} have shown that models based on deep learning architectures can perform similarly to dermatologists on various classification tasks.

However, these ML algorithms are often developed with curated image datasets containing high quality clinical and dermoscopy photographs with limited skin variability, \emph{i.e.}, majority Caucasian or Asian sets in the ISIC dataset (dermoscopy), Asan dataset, Hallym dataset, MED-NODE, Edinburgh dataset \cite{han2018classification}.
The use of such algorithms trained on images often acquired from high quality cameras and/or dermatoscopes may be limited to specialty healthcare facilities and research settings, with questionable transmissibility in resource-limited settings and the primary care, thus creating a gap between healthcare providers and patients.
Smartphone-based imaging is a promising image capture platform for bridging this gap and offering several advantages including portability, cost-effectiveness and connectivity to electronic medical records for secure image transfer and storage. To democratize screening and triage in primary care setting, an ideal ML-based CDS tool should be trained, validated and tested on smartphone-acquired clinical and dermoscopy images, representative of the clinical setting and patient populations for the greatest usability and validity.

While there are challenges to consumer grade smartphone image quality such as variability in angles, lighting, distance from lesion of interest and blurriness, they show promise to improve clinical workflows.
Herein, we propose a two-stage approach to detect skin lesions of interest in wide-field images taken from consumer grade smartphone devices, followed by binary lesion classification into two groups: Malignant \emph{vs.} Benign, for all skin cancers (melanoma, basal cell carcinoma and squamous cell carcinoma) and most common benign tumors.
Ground truth malignancy was ascertained via biopsy, as apposed to consensus adjudication.
As a result, the proposed approach can be integrated and generalized into primary care and dermatology clinical workflows.
Importantly, our work also differs from existing approaches in that our framework can detect lesions from both wide-field clinical and dermoscopy images acquired with smartphones.
%, as opposed to digital cameras as is often the case in existing works.

This paper is organized as follows: in Section~\ref{sc:formulation} we present the problem formulation and the proposed approach.
In Section~\ref{sc:experiments} we describe the data used, the implementation details and quantitative and qualitative experimental results.
Finally, in Section~\ref{sc:discussion} we conclude with a discussion of the proposed approach and acknowledge some limitations of the study.

\begin{figure*}[t!]
    \centering
    \includegraphics[scale=0.285]{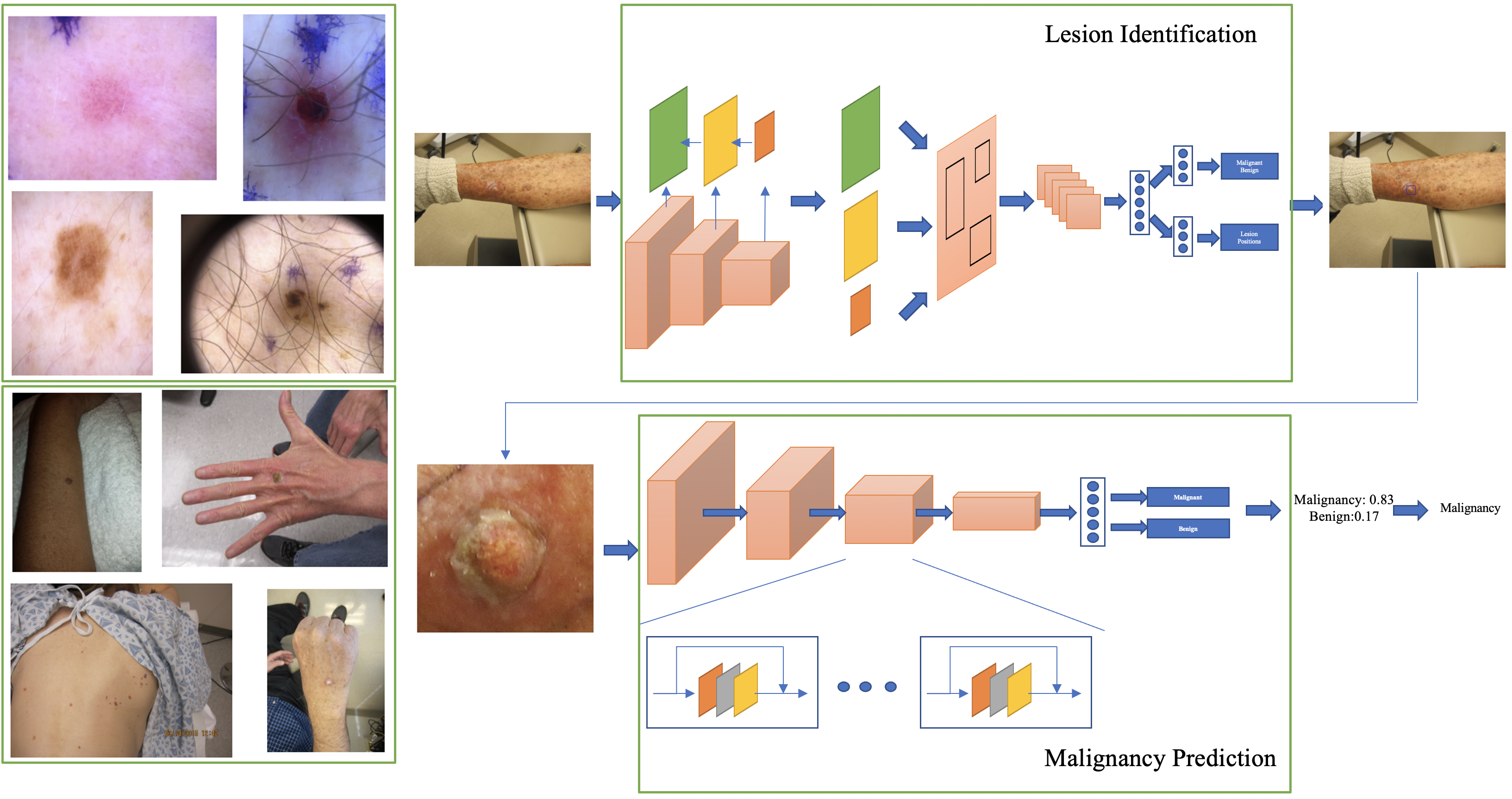}
    \caption{Two-stage malignancy prediction and lesion identification Framework. Top left: Examples of dermoscopy images. Bottom left: Examples of wide-field images. Top right: The lesion identification model estimates lesion locations (bounding boxes) from whole images (dermoscopy or wide-field) via a faster-RCNN architecture (see Section~\ref{sc:identification}). Bottom right: The malignancy prediction model specified via a ResNet-50 architecture predicts the likelihood that a lesion is malignant (see Section~\ref{sc:prediction}). The lesions identified by the lesion identification model are fed into the malignancy prediction model for end-to-end processing.}
    \label{fig:framework}
\end{figure*}

\section{Problem Formulation}\label{sc:formulation}
We represent a set of annotated images as ${\cal D}=\{X_n,Z_n,U_n,y_n\}_{n=1}^N$, where $N$ is the number of instances in the dataset, $X_n \in \mathbb{R}^{h \times w \times 3}$ denotes a color (RBG) image of size $w\times h$ (width $\times$ height) pixels, $Z_n$ is a non-empty set of annotations $Z_n = \{z_{n1},\dots, z_{nm_n}\}$, with elements $z_{ni}$ corresponding to the $i$-th region of interest (ROI) represented as a bounding box with coordinates $(x_{ni}, y_{ni}, w_{ni}, h_{ni})$ (horizontal center, vertical center, width, height) and ROI labels $U_n=\{u_{n1},\ldots,u_{nm_n}\}$, where $m_n$ is the number of ROIs in image $X_n$.
Further, $y_n\in\{0,1\}$ is used to indicate the global image label.

In our specific use case, the images in ${\cal D}$ are a combination of smartphone-acquired wide-field and dermoscopy images with ROIs of 8 different biopsy-confirmed lesion types (ROI labels): Melanoma, Melanocytic Nevus, Basal Cell Carcinoma, Actinic Keratosis/Bowen’s Disease, Benign Keratosis, Dermatofibroma, Vascular Lesions and Other Benign lesions.
The location of different lesions was obtained by manual annotation as described below in Section~\ref{sc:dataset}.
For malignancy prediction, the set of malignant lesions denoted as ${\cal M}$ is defined as Melanoma, Basal Cell Carcinoma, and Actinic Keratosis/Bowen’s Disease/Squamous cell carcinoma while the set of benign lesions contains all the other lesion types.
For the global image label $y_n$, a whole image (smartphone or dermoscopy) is deemed as malignant if at least one of its ROI labels are in the malignant set, ${\cal M}$.

Below, we introduce deep-learning-based models for malignancy prediction, lesion identification and image-level classification for end-to-end processing.
An illustration of the two-step malignancy prediction and lesion identification framework is presented in Figure~\ref{fig:framework}.

\subsection{Malignancy Prediction}\label{sc:prediction}
Assuming we know the position of the ROIs, \emph{i.e.}, $\{X_n,Z_n\}_{n=1}^N$ are always available, the problem of predicting whether a lesion is malignant can be formulated as a binary classification task.
Specifically, we specify a function $f_\theta(\cdot)$ parameterized by $\theta$ whose output is the probability that a single lesion is consistent with a malignancy pathohistological finding in the area, \emph{i.e.},
\begin{align}\label{eq:prediction}
    p(u_{ni}\in{\cal M}|X_n, z_{ni})=f_\theta(X_n, z_{ni}) ,
\end{align}
where $f_\theta(\cdot)$ is a convolutional neural network that takes the region of $X_n$ defined in $z_{ni}$ as input.
In practice, we use a ResNet-50 architecture \cite{he2016deep} with additional details described in Section~\ref{sc:model_details}.

\subsection{Lesion identification}\label{sc:identification}
Above we assume that the location (ROI) of the lesions is known, which may be the case in dermoscopy images as illustrated in Figure~\ref{fig:framework}.
However, in general, wide-field dermatology images are likely to contain multiple lesions, while their locations are not known or recorded as part of clinical practice.
Fortunately, if lesion locations are available for a set of images (via manual annotation), the task can be formulated as a supervised object detection problem, in which the model takes the whole image as input and outputs a collection of predicted ROIs along with their likelihood of belonging to a specific group.
Formally,
\begin{align}\label{eq:identification}
    \left\{\hat{z}_{ni}, \hat{p}_{ni}, \right\}_{i = 1}^{\hat{m}_{n}} = g_\psi(X_n) ,
\end{align}
where $\hat{p}_{ni}=[\hat{p}_{ni1},\ldots,\hat{p}_{niC}]\in(0,1)^C$ is the likelihood that the predicted region $\hat{z}_{ni}=\{\hat{x}_{ni}, \hat{y}_{ni}, \hat{w}_{ni}, \hat{h}_{ni}\}$ belongs to one of $C$ groups of interest, \emph{i.e.}, $p(\hat{z}_{ni}\in c)= \hat{p}_{nic}$.
In our case, we consider three possible choices for $C$, namely, $i$) $C=1$ denoted as \emph{one-class} where the model seeks to identify any lesion regardless of type; $i$) $C=2$ denoted as \emph{malignancy} in which the model seeks to separately identify malignant and benign lesions; and $iii$) $C=8$ denoted as \emph{sub-type}, thus the model is aware of all lesion types of interest.

Note that we are mainly interested in finding malignant lesions among all lesions present in an image as opposed to identifying the type of all lesions in the image.
Nevertheless, it may be beneficial for the model to be aware that different types of lesions may have common characteristics which may be leveraged for improved detection.
Alternatively, provided that some lesion types are substantially rarer than others (\emph{e.g.}, dermatofibroma and vascular lesions only constitutes 1\% each of all the lesions in the dataset described in Section~\ref{sc:dataset}), seeking to identity all lesion types may be detrimental for the overall detection performance.
This label granularity trade-off will be explored in the experiments.
In practice, we use a Faster-RCNN (region-based convolutional neural network) \cite{NIPS2015_14bfa6bb} with a feature pyramid network (FPN) \cite{lin2017feature} and a ResNet-50 \cite{he2016deep} backbone as object detection architecture.
Implementation details can be found in Section~\ref{sc:model_details}.

\subsection{Image classification}\label{sc:classification}
For screening purposes, one may be interested in estimating whether an image is likely to contain a malignant lesion so the case can be directed to the appropriate dermatology specialist.
In such case, the task can be formulated as a whole-image classification problem
\begin{align}\label{eq:classification}
    p( y_n = 1 | X_n) = h_\phi( X_n ) , 
\end{align}
where $p(y_n=1|X_n)\in(0,1)$ is the likelihood that image $X_n$ contains a malignant lesion.

The model in~\eqref{eq:classification} can be implemented in a variety of different ways.
Here we consider three options, two of which leverage the malignancy prediction and lesion identification models described above.

\paragraph{Direct image-level classification} $h_\phi(\cdot)$ is specified as a convolutional neural network, \emph{e.g.}, ResNet-50 \cite{he2016deep} in our experiments, to which the whole image $X_n$ is fed as input.
Though this is a very simple model that has advantages from an implementation perspective, it lacks the context provided by (likely) ROIs that will make it less susceptible to interference from background non-informative variation, thus negatively impacting classification performance.

\paragraph{Two-stage approach} $h_\phi(\cdot)$ is specified as the combination of the \emph{one-class} lesion identification and the malignancy prediction models, in which detected lesions are assigned a likelihood of malignancy using~\eqref{eq:prediction}.
This is illustrated in Figure~\ref{fig:framework}(Right).
Then we obtain
\begin{align}
    p( y_n = 1 | X_n)=a\left( \{p(u_{ni}\in{\cal M}|X_n, \hat{z}_{ni})\}_{i=1}^{\hat{m}_n} \right) ,
\end{align}
where we have replaced the ground truth location $z_{ni}$ in \eqref{eq:prediction} with the $\hat{m}_n$ predicted locations from \eqref{eq:identification}, and $a(\cdot)$ is a permutation-invariant aggregation function.
In the experiments we consider two simple parameter-free options:
\begin{align}
    a(\cdot)= & \frac{1}{\hat{m}_n}\sum_{i=1}^{\hat{m}_n} p(u_{ni}\in{\cal M}|X_n, \hat{z}_{ni}) , \ \ & {\rm (Average)} \label{eq:avg} \\
    a(\cdot)= & \max(\{p(u_{ni}\in{\cal M}|X_n, \hat{z}_{ni})\}_{i=1}^{\hat{m}_n}) & {\rm (Maximum)} \label{eq:max} \\
    a(\cdot)= & 1 - \prod_{i=1}^{\hat{m}_n} p(u_{ni}\in{\cal M}|X_n, \hat{z}_{ni}) \ \ & \text{(Noisy OR)} \label{eq:nor}
\end{align}
Other more sophisticated (parametric) options such as noisy AND \cite{kraus2016classifying}, and attention mechanisms \cite{NIPS2017_3f5ee243}, may further improve performance but are left as interesting future work.

\paragraph{One-step approach} $h_\phi(\cdot)$ is specified directly from the \emph{sub-types} lesion identification model in \eqref{eq:identification} as
\begin{align}
    p( y_n = 1 | X_n)=a\left( \{\hat{p}_{ni}\}_{i=1}^{\hat{m}_n} \right) ,
\end{align}
where $a(\cdot)$ is either \eqref{eq:avg}, \eqref{eq:max} or \eqref{eq:nor}.

From the options described above, the direct image-level classification approach is conceptually simpler and easier to implement but it does not provide explanation (lesion locations) to its predictions.
The one-step approach is a more principled end-to-end system that directly estimates lesion locations, lesion sub-type likelihood, and overall likelihood of malignancy, however, it may not be suitable in situations where the availability of labeled sub-type lesions may be limited, in which case, one may also consider replacing the \emph{sub-type} detection model with the simpler \emph{malignancy} detection model.
Akin to this simplified one-step approach, the two-stage approach provides a balanced trade-off between the ability of estimating the location of the lesions and the need to identify lesion sub-types.
All these options will be quantitatively compared in the experiments below.

\section{Experiments}\label{sc:experiments}
Comprehensive experiments to analyze the performance of the proposed approach were performed. 
First, we describe the details of the dataset and the models being considered, and present evaluation metrics for each task, while comparing various design choices described in the previous section.
Then, we study the effects of adding clinical covariates and using an auxiliary publicly available dataset for data augmentation.
Lastly, we present some visualization of the proposed model predictions for qualitative analysis.

\begin{table}[t!]
    \centering
    \caption{Lesion type counts by dataset. The 8 lesion types considered are: Melanoma (MEL), Melanocytic Nevus (NV), Basal Cell Carcinoma (BCC), Actinic Keratosis/Bowen’s Disease (AKIEC), Benign Keratosis (BKL), Dermatofibroma (DF), Vascular Lesions (VASC) and Other Benign (OB) lesions.}
    \label{tab:lesion_counts}
    \adjustbox{tabular=lrrr,width=\columnwidth}{
        Lesion Type & Discovery & ISIC2018 & Test \\
        \hline
        MEL & 596 (7\%) & 1,113 (11\%) & 50 (10\%) \\
        NV & 1,343 (16\%) & 6,705 (67\%) & 139 (27\%) \\
        BCC & 1,627 (20\%) & 514 (5\%) & 76 (15\%) \\
        AKIEC & 2,473 (30\%) & 327 (3\%) & 121 (24\%) \\
        BKL & 974 (12\%) & 1,099 (11\%) & 83 (16\%) \\
        DF & 97 (1\%) & 115 (1\%) & 11 (2\%) \\
        VASC & 106 (1\%) & 142 (2\%) & 5 (1\%) \\
        OB & 1,027 (13\%) & - & 26 (5\%) \\
    }
\end{table}

\subsection{Dataset}\label{sc:dataset}
\paragraph{Discovery Dataset}
To develop the model we consider a single institution, retrospective collection of skin lesion images taken with smartphones with and without dermoscopy from Duke University Medical Center patients of age 18 and older from 2013-2018.
The \emph{discovery} dataset consists of 7,752 images from 4,185 patients with 8,243 manually annotated lesions, from which 4,696 (57\%) lesions in 3,425 images are malignant.
In terms of skin tone, 6,529 images (6,947 lesions) are light, 1,118 images (1,183 lesions) are medium and 105 images (113 lesions) are dark tone.
Lesions were manually annotated as bounding boxes (ROIs) by a dermatology trained medical doctor (Dr. Kheterpal, MK) using a in-house annotation application.
Diagnoses taken from the biopsy reports associated with the lesion images were designated as the ground truth (Malignant \emph{vs.} Benign).
Further, there are 859 (11\%) dermoscopy images and 6,893 (89\%) wide-field images.
Table~\ref{tab:lesion_counts} shows detailed lesion type counts and proportions.
The average area of the lesion is 14,431 (Q1-Q3: 3,834-88,371) pixels$^2$ (roughly $120\times 120$ pixels in size) while the average area of the images is 7'200,000 (3'145,728-12'000,000) pixels$^2$ (roughly $2683\times 2683$ pixels in size).
We split the dataset, at the patient level, into 7,002 lesions (6,593 images) for training and 1,241 lesions (1,159 images) for validation.
The validation set was used to optimize the model parameters, architecture and optimization parameters.

\paragraph{Clinical Dataset}
We also consider a subset of 2,464 images from 2,149 patients for which we also have demographic (age at encounter, sex and race), lesion characteristics (location and number of previous dermatology visits), comorbidities (history of chronic ulcer of skin, diseases of white blood cells, human immunodeficiency virus infection, Hodgkin’s disease, non-Hodgkin’s lymphoma, infective arthritis and osteomyelitis, leukemias, Parkinson’s disease, rheumatologic diseases, skin and subcutaneous tissue infections, inflammatory condition of skin, systemic lupus erythematosus, other connective tissue disease, other sexually transmitted diseases, other hematologic diseases, and other skin disorders) and skin-cancer-related medications (immunosuppressants, corticosteroids, antihypertensives, antifungals, diuretics, antibiotics, antiarrhythmics, antithrombotics, chemotherapy, targeted therapy, immunotherapy, and other), their risk (Low \emph{vs.} High), and frequency of administration.
Among these patients, 1311 (1503 images) are diagnosed as malignant and 838 (961 images) as benign.
Similar to the discovery dataset, we split these data into 85\% for training and the remaining 15\% for validation.

\paragraph{ISIC2018}
Provided that we have a smaller number of dermoscopy images, we also consider augmenting our discovery dataset with the ISIC2018 training dataset \cite{codella2019skin,tschandl2018ham10000} consisting of 10,015 dermoscopy images, from which 1,954 correspond to malignant lesions and 8,061 benign lesions. Detailed lesion type counts are presented in Table 1.
In the experiments, we also consider the ISIC2018 validation dataset to test the model with and without ISIC2018 augmentation.

\paragraph{Independent Test Set}
In order to evaluate the performance of the model relative to human diagnosis, we consider an independent set of 497 images also from Duke University Medical Center patients.
In terms of skin tone, 366 images (376 lesions) are light, 120 images (124 lesions) are medium and 11 images (11 lesions) are dark.
From these images, 242 are malignant and 255 are benign.
To compare the proposed model with human experts, we had three dermatology trained medical doctors with different levels of experience label each of the images without access to the biopsy report or context from the medical record.
In terms of experience, MJ has 3 years dermoscopy experience, AS has 6 years of dermoscopy experience and MK has 10 year dermoscopy experience.
Provided that MK also participated in lesion annotation with access to biopsy report information, we allowed 12 months separation between the lesion annotation and malignancy adjudication sessions.
Detailed lesion type counts are presented in Table~\ref{tab:lesion_counts}.
The average area of the lesion is 186,276 (7,379-153,169) pixels$^2$ (roughly $432\times 432$ pixels in size) while the average area of the images is 7'200,000 (3'145,728-12'000,000) unbelievable it is same as train but it is true) pixels$^2$ (roughly $2683\times 2683$ pixels in size).

\subsection{Model Details}\label{sc:model_details}
\paragraph{Malignancy Classification}
For malignancy classification we use a ResNet-50 architecture \cite{he2016deep} as shown in Figure~\ref{fig:framework}(Bottom right).
The feature maps obtained from the last convolutional block are aggregated via average pooling and then fed through a fully connected layer with sigmoid activation that produces the likelihood of malignancy.
The model was initialized from a ResNet-50 pre-trained on ImageNet and then trained (refined) using a stochastic gradient descent (SGD) optimizer for 100 epochs, with batch size 64 initial learning rate 0.01, momentum 0.9 and weight decay 1e-4.
The learning rate was decayed using a half-period cosine function, \emph{i.e.}, $\eta \left( t \right) = 0.01 \times \left[ 0.5 + 0.5 \cos \left( t \pi / T_{\text{max}} \right) \right]$, where $t$ and $T_{\text{max}}$ are the current step and the max step, respectively.
%
% For malignancy classification, we use a ResNet-50, whose architecture details are shown as ``Classification' in Figure (Framework) The feature from the last layer is global average pooled, then mapped to the malignancy score using a fully connected layer and a sigmoid non-linearity. The model was trained using SGD optimizer for 100 epochs, with batch size 64 initial learning rate 0.01, momentum 0.9 and weight decay 1e-4. The learning rate was decayed using a half-period cosine function, i.e. $\eta \left( t \right) = 0.01 \times \left[ 0.5 + 0.5 \cos \left( t \pi / T_{\text{max}} \right) \right]$, where $t$ and $T_{\text{max}}$ is the current step and the max step, respectively.

\paragraph{Lesion Identification}
The lesion identification model is specified as a Faster-RCNN \cite{NIPS2015_14bfa6bb} with a FPN \cite{lin2017feature} and a ResNet-50 \cite{he2016deep} backbone.
%The bottom-up part is identical to ResNet-50, and every feature map in the top-down part is a fusion of upsampled feature map from previous stage and a feature map with the same resolution from the bottom-up part.
%
The feature extraction module is a ResNet-50 truncated to the 4-th block.
The FPN then reconstructs the features to higher resolutions for better multi-scale detection \cite{lin2017feature}.
Higher resolution feature maps are built as a combination of the same-resolution ResNet-50 feature map and the next lower-resolution feature map from the FPN, as illustrated in Figure~\ref{fig:framework}(Top right).
The combination of feature maps from the last layer of the feature extraction module and all feature maps from the FPN are then used for region proposal and ROI pooling.
See \cite{lin2017feature} for further details.
% is built by fusion of the previous, lower resolution feature map and a same-resolution feature map from the feature extraction part.
% The last level feature produced by feature extraction and all features generated by FPN are all used for region proposal and RoI pooling.
% \rh{this needs to be explained better: what is bottom-up, what is top-down? what is the previous stage?}
The model was trained using an SGD optimizer for 80,000 steps, with batch size of 512 images, initial learning rate 0.001, momentum 0.9 and weight decay 1e-4.
Learning rate was decayed 10x at 60,000-th and 80,000-th step, respectively.

\paragraph{Direct Image-Level Classification Model}
The direct image-level classification model in Section~\ref{sc:classification} has the same architecture and optimization parameters as the malignancy classification model described above.

\paragraph{Clinical Model}
The clinical model was built using logistic regression with standardized input covariates and discrete (categorical) covariates encoded as one-hot-vectors. 
% The clinical record logistic regression model was trained using scikit-learn. We use the default parameters, expect that we extend the max step to 200 for better convergence. Every feature dimension was normalized to zero mean and unit variance before feeding into the model.

\paragraph{Combined Model}
% \rh{Need to provide the details of the combined model (clinical + images).}
In order to combine the clinical covariates with the images into a single model, we use the malignancy classification model as the backbone while freezing all convolutional layers during training.
Then, we concatenate the standardized input covariates and the global average-pooled convolutional feature maps, and feed them through a fully connected layer with sigmoid activation that produces the likelihood of malignancy.
% For combining the clinical records with images, we choose to use malignancy classification model as the backbone and freeze its all convolution layers. We concatenate the normalized clinical record and the global average pooled CNN feature. A new fully connected layer is trained to do the prediction using both features.
The combined model was trained using an SGD optimizer for 30 epochs, with batch size 64, initial learning rate 0.001, momentum 0.9 and weight decay 1e-4.
The learning rate was decayed using a half-period cosine function as in the malignancy classification model.
%, which is same as the malignancy classification model did.
%
% Unless otherwise specified, we use the validation dataset for model selection, \emph{i.e.}, we report the \textit{test} subset performance of the best model selected by \textit{validation} subset performance.

\paragraph{Implementation}
We used Detectron \cite{Detectron2018} for the lesion identification model.
All other models were coded in Pyhton 3.6.3 using the PyTorch 1.3.0 framework except for the clinical model that was implemented using scikit-learn 0.19.1.
The source code for all the models used in the experiments is available (upon publication) at \url{github.com/user/dummy}.

\subsection{Performance Metrics}\label{sc:performance}
For malignancy prediction, two threshold-free metrics of performance are reported, namely, area under the curve (AUC) of the receiving operating characteristic (ROC) and the average precision (AP) of the precision recall curve, both described below.
AUC is calculated as:
\begin{align*}
    \text{AUC} & = \frac{1}{2}\sum_{i} \left[ \text{FPR}_{\hat{p}_{i+1}} - \text{FPR}_{\hat{p}_{i}} \right]
    \left[ \text{TPR}_{\hat{p}_{i+1}} + \text{TPR}_{\hat{p}_{i}} \right] \\
    \text{TPR}_t & = p( \hat{p} > t | y = 1) \\
    \text{FPR}_t & = p( \hat{p} > t | y = 0),
\end{align*}
where $t\in[\hat{p}_1,\ldots,\hat{p}_i,\hat{p}_{i+1},\ldots]$ is a threshold that takes values in the set of sorted test predictions $\{\hat{p}_i\}_{i=1}^N$ from the model, and the true positive rate, TPR$_t$, and false positive rate, FPR$_t$, are estimated as sample averages for a given threshold $t$.

Similarly, the AP is calculated as:
\begin{align*}
    \text{AP} & = \frac{1}{2}\sum_{i} \left[ TPR_{\hat{p}_{i+1}} - TPR_{\hat{p}_{i}} \right]\left[ \text{PPV}_{\hat{p}_{i+1}} + \text{PPV}_{\hat{p}_{i}} \right] \\
    \text{PPV}_{t} & = p( y = 1 | \hat{p} > t ) ,
\end{align*}
where PPV$_t$ is the positive predictive value or \emph{precision} for threshold $t$.
The calculation for the AUC and AP areas follow the trapezoid rule. 

The intersection over union (IoU) is defined as the ratio between the overlap or ground truth and estimated ROIs, $\{z_{ni}\}_{i=1}^{m_n}$ and $\{\hat{z}_{ni}\}_{i=1}^{\hat{m}_n}$, respectively, and the union of their areas.
For a given ROI, IoU=1 indicates complete overlap between prediction and ground truth.
Alternatively, IoU=0 indicates no overlap.
In the experiments, we report the median and interquartile range IoU for all predictions in the test set.

The mean average precision (mAP) is the AP calculated on the binarized predictions from the detection model such that predictions with an IoU$\geq t$ are counted as correct predictions or incorrect otherwise, if IoU$< t$, for a given IoU threshold $t$ set to 0.5, 0.75 and (0.5,0.95) in the experiments.
These values are standard in object detection benchmarks, see for instance \cite{lin2014microsoft}.

We also report the recall with IoU$>0$ as a general, easy to interpret, metric of the ability of the model to correctly identify lesions in the dataset.
Specifically, we calculate it as the proportion of lesions (of any type) in the dataset for which predictions overlap with the ground truth.

\subsection{Quantitative Results}
\paragraph{Malignancy Prediction}
First, we present results for the malignancy prediction task, for which we assume that lesions in the form of bounding boxes (ROIs) have been pre-identified from smartphone (wide-field) or dermoscopy images.
Specifically, we use ground truth lesions extracted from larger images using manual annotations as previously described. 
Table~\ref{tab:malignancy_prediction} shows AUCs and APs for the malignancy prediction model described in Section~\ref{sc:prediction} on the independent test dataset.
We observe that the model performs slightly better on dermoscopy images presumably due to their higher quality and resolution.

\begin{table}[t!]
    \centering
    \caption{Malignancy prediction from ground truth lesions manually annotated as bounding boxes a dermatology trained medical doctor (MK).}
    \label{tab:malignancy_prediction}
    \adjustbox{tabular=lcc,width=\columnwidth}{
    %\hline
    & AUC & AP \\
    \hline
    All lesions & 0.779 & 0.752 \\
    %\hline
    Lesions from smartphone images & 0.779 & 0.749 \\
    %\hline
    Lesions from dermoscopy images & 0.787 & 0.760 \\
    %\hline
    }
\end{table}

\begin{figure*}[t!]
%\begin{subfigure}{.31\textwidth}
  \centering
  \includegraphics[width=.33\linewidth]{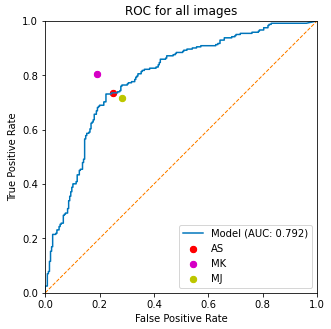}
  %\caption{1a}
  %\label{fig:sfig3}
%\end{subfigure}%
%\begin{subfigure}{.31\textwidth}
  %\centering
  \includegraphics[width=.33\linewidth]{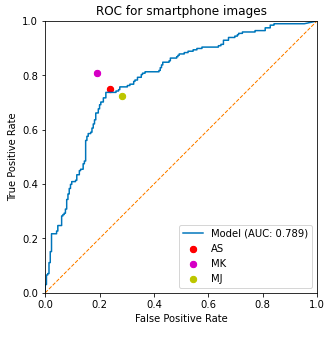}
  %\caption{1b}
  %\label{fig:sfig4}
%\end{subfigure}
%\begin{subfigure}{.31\textwidth}
  %\centering
  \includegraphics[width=.33\linewidth]{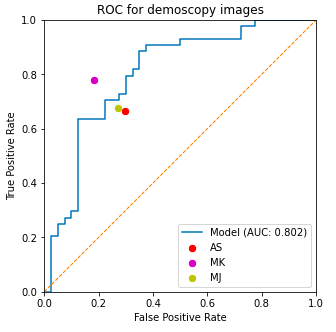}
  %\caption{1c}
  %\label{fig:sfig4}
%\end{subfigure}\\
%\begin{subfigure}{.31\textwidth}
  %\centering
  \includegraphics[width=.33\linewidth]{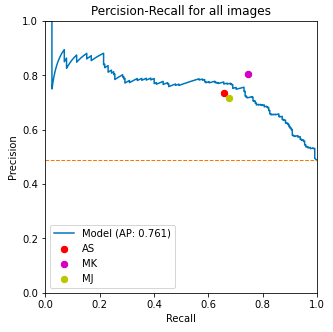}
  %\caption{1d}
  %\label{fig:sfig3}
%\end{subfigure}%
%\begin{subfigure}{.31\textwidth}
  %\centering
  \includegraphics[width=.33\linewidth]{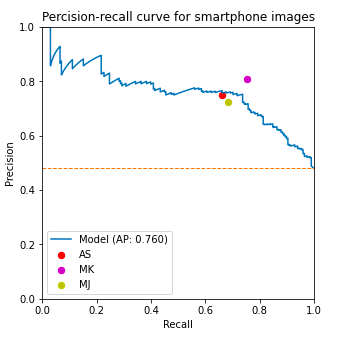}
  %\caption{1e}
  %\label{fig:sfig4}
%\end{subfigure}
%\begin{subfigure}{.31\textwidth}
  %\centering
  \includegraphics[width=.33\linewidth]{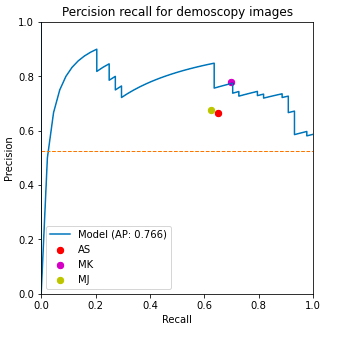}
  %\caption{1f}
  %\label{fig:sfig4}
%\end{subfigure}
\caption{Performance metrics of the malignancy prediction models. ROC and PR curves, top and bottom rows, respectively, for all images (Left), smartphone (wide-field) images (Middle) and dermoscopy images (Right) on the test set. Predictions were obtained from the \emph{one-class} model followed by the malignancy prediction model and the image classification aggregation approach. Also reported are the TPR (sensitivity) and FPR (1-specificity) for three dermatology trained MDs (AS, MK and MJ).}
\label{fig:malignancy_image_res}
\end{figure*}

\paragraph{Malignancy Detection}
Provided that in practice lesions are not likely to be pre-identified by clinicians, we present automatic detection (localization) results using the models presented in Section~\ref{sc:identification}.
Specifically, we consider three scenarios: $i$) \emph{one-class}: for all types of lesions combined; $ii$) \emph{malignancy}: for all types of lesions combined into malignant and benign; and $iii$) \emph{sub-type}: for all types of lesions separately.
Table~\ref{tab:detection_map} shows mean Average Precision (mAP) at different thresholds, Recall (sensitivity) and IoU summaries (median and interquartile range), all on the independent test set.
In order to make mAP comparable across different scenarios, we calculate it for all lesions regardless of type, \emph{i.e.}, mAP is not calculated for each lesion type and then averaged but rather by treating all predictions as lesions. 
We observe that in general terms, the \emph{one-class} lesion identification model outperforms the more granular \emph{malignancy} and \emph{sub-type} approaches.
These observation is also consistent in terms of Recall and IoU.

For the \emph{one-class} model specifically, 82.9\% regions predicted are true lesions at at IoU$\geq 0.5$ (at least 50\% overlap with ground truth lesions), whereas the precision drops to 26.8\% with a more stringent IOU$\geq 0.75$.
Interestingly, the 95.6\% Recall indicates that the \emph{one-class} model is able to capture most of the true lesions at IoU$> 0$ and at least 50\% of the predicted regions have a IoU$>0.73$ or IoU$>0.59$ for 75\% of the lesions in the independent test set.

% \rh{we need to say something about these results, are they good? which one is better? what is the purpose of using different thresholds? We also need to show interpretable results: what proportion of the test set lesions can we capture using the model? can we also present intersection over union results?} 
% We then evaluate the lesion identification models based on the traditional object detection metric: mean Average Precision (mAP). 
% The results are showed as table \ref{tab:detection_map}, where 'one class' represents the model just identify the lesions area, the 'binary classes' and 'eight classes' mean the model detect the specific classes lesions: 'binary classes' for malignant and benign, 'eight classes' for the mentioned 8 classes.

\begin{table}[t!]
    \centering
    \caption{Lesion detection from smartphone (wide-field) and dermoscopy images. Performance is evaluated as the mean Average Precision (mAP) at three different thresholds: $0.5$, $0.75$ and $[0.5,0.95]$, recall (sensitivity) and intersection over union (IoU) summarized as median (interquartile range).}
    \label{tab:detection_map}
    \adjustbox{tabular=lccc,width=\columnwidth}{
    %\hline
    %\hline
        & One-class & Malignancy & Sub-type \\
    \hline
    mAP$_{@0.5}$ & {\bf 0.829} & 0.807 &   0.762\\
    %\hline
    mAP$_{@0.75}$ & {\bf 0.268} & 0.245 &  0.256\\
    %\hline
    mAP$_{@0.5,0.95}$ & {\bf 0.380} & 0.361 & 0.352 \\
    Recall & {\bf 0.959} & 0.9530 & 0.9354 \\
    IoU & $\bm{0.73_{(0.59,0.82)}}$ & $0.71_{(0.59,0.81)}$ & $0.71_{(0.61, 0.80)}$ \\
    %\hline
    }
\end{table}

\paragraph{Image Classification}
The image-level prediction results of malignancy are reported in Figure \ref{fig:malignancy_image_res}.
Predictions on the independent test set were obtained from the average-pooled image classification model in Section~\ref{sc:classification} with the \emph{one-class} detection model in Section~\ref{sc:identification} and the malignancy prediction model in Section~\ref{sc:prediction}.
From the performance metrics reported we note that the proposed approach is comparable with manual classification by three expert dermatologists (AS, MK and MJ).
Interestingly, in dermoscopy images, the model slightly outperforms two of the three dermatologists and the difference in their performance is consistent with their years of experience; MK being the most experienced and better performing dermatologist.

Additional results comparing the different image-level malignancy prediction strategies described in Section~\ref{sc:classification}, namely, $i$) direct image-level classification, $ii$) two-stage with \emph{one-class} lesion identification, and one-step with $iii$) \emph{malignancy} or $iv$) \emph{sub-type} identification models with average pooling aggregation are presented in Table~\ref{tab:image_cls_rest_method_Benign}.
In terms of AUC, the one-class approach consistently outperforms the others, while in terms of AP, sub-type is slightly better.
Interestingly, the direct image-level classification which takes the whole image as input, without attempting to identify the lesions, performs reasonably well and may be considered in scenarios where computational resources are limited, \emph{e.g.}, mobile and edge devices.

\begin{table}[t!]
    \centering
    \caption{Performance metrics with different image-level classification strategies (direct image-level, two-stage with \emph{one-class} lesion identification and one-step with \emph{malignancy} or \emph{sub-type} identification) stratified into all images and smartphone or dermoscopy only subsets.}
    \label{tab:image_cls_rest_method_Benign}
    \adjustbox{tabular=lcccccc,width=\columnwidth}{
    %\hline
    & \multicolumn{2}{c}{All Images} &
    \multicolumn{2}{c}{Smartphone Only} & \multicolumn{2}{c}{Dermoscopy Only} \\
    \hline
    & AUC & AP & AUC & AP & AUC & AP \\
    %\hline
    \hline
    Image-level & 0.742 & 0.736 & 0.735 & 0.725 & 0.768 & 0.791 \\
    %\hline
    Malignancy  & 0.763 & 0.745 & 0.772 & 0.748 & 0.699 & 0.759 \\
    %\hline
    Sub-type & 0.777 & {\bf 0.778} & 0.787 & {\bf 0.783} & 0.739 & {\bf 0.789} \\
    %\hline
    One-class & {\bf 0.792} & 0.761 & {\bf 0.789} & 0.759 & {\bf 0.802} & 0.766 \\
    %\hline
    %\end{tabular}
    }
\end{table}

Further, we also compare different aggregation strategies (average, max and noisy OR pooling) described in Section~\ref{sc:classification} and lesion identification models (\emph{one-class}, \emph{malignancy} and \emph{sub-types}) described in Section~\ref{sc:identification} are presented in Table~\ref{tab:pooling}, from which we see that the combination of average pooling and one-class lesion detection slightly outperforms the alternatives.

\begin{table}[t!]
    \centering
    \caption{Performance metrics of the image-level malignancy prediction model with different aggregation strategies (average, noisy OR and max pooling) and lesion identification models (\emph{one-class}, \emph{malignancy} and \emph{sub-types}). The best performing combination is highlighted in boldface.}
    \label{tab:pooling}
    \adjustbox{tabular=lcccccc,width=\columnwidth}{
    %\hline
    & \multicolumn{2}{c}{Average} & 
    \multicolumn{2}{c}{Noisy OR} &
    \multicolumn{2}{c}{Max}\\
    %\hline
    Detection & AUC & AP & AUC & AP & AUC & AP \\
    \hline
    Malignancy & 0.7627 & 0.7447 & 0.7644 & 0.7464 & 0.7463 & 0.7343 \\
    %\hline
    Sub-type & 0.7771 & 0.7775 & 0.7769 & 0.7698 & 0.7715 & 0.7774 \\
    %\hline
    Two-stage & {\bf 0.7924} & {\bf 0.7607} & 0.7826 & 0.7653 & 0.7825 & 0.7631 \\
    %\hline
    }
\end{table}

% One thing need to be noticed here that each image may includes multiple lesions, which means  we need to combine all lesions results of an image as the binary prediction of this image. We tried two combination methods: for one image, the final score can be  the maximum predicted lesion score in it or the averaging of all predicted lesions scores in it. Since the averaging method is fairer, which takes all lesions predictions of one image into account, we show the performance based on it (the results for maximum method are provided in Appendix).
%
% Both the lesion identification models and the two-stage pipeline may detect and classify multiple lesions for one image, so we calculate their image predictions by applying the same averaging method we used for malignancy prediction model. And because the image classification task is a binary task, we convert the predict scores of lesions of eight-class detection model from 8 classes scores into 2 classes scores: for each lesion, sum the predict scores of malignant classes (classes 0,2,3) as the malignancy scores and sum the rest as the benign scores. And if the detection model do not detect any lesion for one image, we will assume the prediction of this image will be malignancy (the worst assumption).The image prediction results of different models are reported as table \ref{tab:image_cls_rest_meth}.

\begin{figure*}[t!]
%\begin{subfigure}{.5\textwidth}
  \centering
  \includegraphics[width=.45\linewidth]{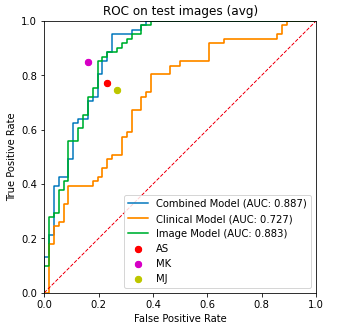}
  %\caption{2a}
  %\label{fig:sfig3}
%\end{subfigure}%
%\begin{subfigure}{.5\textwidth}
  %\centering
  \includegraphics[width=.45\linewidth]{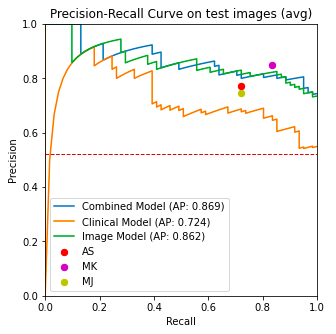}
  %\caption{2b}
  %\label{fig:sfig4}
%\end{subfigure}
\caption{Performance metrics of the malignancy prediction models including clinical covariates. ROC and PR curves for three models are presented, namely, combined (clinical + images), image only and clinical covariates only. Also reported are the TPR (sensitivity) and FPR (1-specificity) for three dermatology trained MDs (AS, MK and MJ).}
\label{fig:clinical records}
\end{figure*}

\begin{table*}[t!]
    \centering
    \caption{Performance metrics (AUC and AP) of the models with data augmentation. We consider three models with and without ISIC2018 dermoscopy image dataset augmentation. The three models considered are the malignancy prediction model described in Section~\ref{sc:prediction}, and the direct image-level classification and two-step approach with one-class lesion identification described in Section~\ref{sc:classification}.}
    \label{tab:ISIC}
    %\begin{adjustbox}{center}
    \begin{tabular}{llcccccc}
    %\hline
    & &\multicolumn{2}{c}{All Images} &
    \multicolumn{2}{c}{Smartphone Only} & \multicolumn{2}{c}{Dermoscopy Only} \\
    %\hline
    & & AUC & AP & AUC & AP & AUC & AP \\
    \hline
    %\hline
    \multirow{2}{*}{Malignancy Prediction} & Discovery & 0.776 & 0.751 & 0.774 & 0.749 & 0.784 & 0.755 \\
    %\hline
    & Discovery + ISIC2018 & 0.787 & 0.770 & 0.793 & 0.771 & 0.762 & 0.781 \\
    \hline
    \multirow{2}{*}{Direct Image-level} & Discovery & 0.742 & 0.736 & 0.735 & 0.725 & 0.768 & 0.791 \\
    %\hline
    & Discovery + ISIC2018 & 0.767 & 0.752 & 0.780 & 0.758 & 0.717 & 0.744 \\
    \hline
    \multirow{2}{*}{Two-step Approach} & Discovery & 0.792 & 0.761 & 0.789 & 0.759 & 0.803 & 0.766 \\
    %\hline
    & Discovery + ISIC2018 & 0.803 & 0.782 & 0.803 & 0.772 & 0.808 & 0.832 \\
    \hline
    \multirow{2}{*}{Malignancy Prediction} & ISIC2018 & - & - & - & - &0.959 & 0.849\\
    %\hline
    & Discovery + ISIC2018 & - & - & - & - & 0.970 & 0.891
    %\hline
    \end{tabular}
    %\end{adjustbox}
\end{table*}

\subsubsection{Accounting for Clinical Data}
Next, we explore the predictive value of clinical features and their combination with image-based models.
Specifically, we consider three models: $i$) the logistic regression model using only clinical covariates; $ii$) the two-stage approach with \emph{one-class} lesion identification; and $iii$) the combined model described in Section~\ref{sc:model_details}.
Note that since we have a reduced set of images for which both clinical covariates and images are available as described in Section~\ref{sc:dataset}, all models have been re-trained accordingly.
Figure~\ref{fig:clinical records} shows ROC and PR curves for the three models and the TPR and FPR values for three dermatology trained MDs on the independent test set.
Results indicate a minimal improvement in classification metrics by combining clinical covariates and images, and a significant improvement of the image-based models relative to the pure clinical model, which underscores the importance and predictive value of the images for the purpose of malignancy prediction.

% In previous work, some other modalities are also utilized for dermatology lesion detection. In this section we experimented with 2 popular choices: clinical records and dermoscopy images, and study whether they can further improve our model.
%
% First we collected the clinical records of each patient in our dataset. We benchmark the effectiveness of clinical records by using different combinations of features: (1) clinical records only; (2) image only and (3) clinical results plus image. As clinical records form a plain vector of numbers, we use them by directly building a logistic regression model. Note that although clinical records are collected at patient level, the model is trained and tested at image level, i.e. if there are multiple images of a single patient, duplicated record-label pairs will exist in the clinical records, and in the rare case that a patient has different type of lesions, contradicting items will appear. This is for fair comparison to our image model. The image model and clinical record model are combined by taking average of their final classification scores. The comparison results are in Figure \ref{fig:clinical records}. Note that in this set of experiments, we eliminated the images for which we are unable to collect clinical records.
%
% From the results we can see that the image model almost dominated the clinical record model: its performance is much better, and the combined model brings no noticeable performance boost compared to the image-only model.

\begin{figure*}[t!]
  \centering
  \includegraphics[width=1\linewidth]{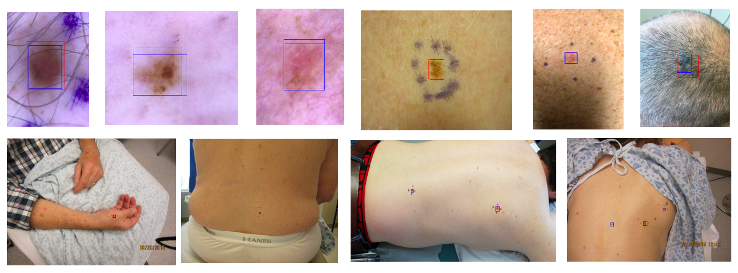}
  \label{fig:sfig3}
\caption{Lesion detection examples. Top: Dermoscopy images. Bottom: Smartphone (wide-field) images. The ground-truth, manually annotated lesion is represented by the red bounding box, while the predicted lesion is denoted by the blue bounding box.}
\label{fig:detection examples}
\end{figure*}

\begin{figure}[t!]
%\begin{subfigure}{.5\textwidth}
  \centering
  \includegraphics[width=.95\columnwidth]{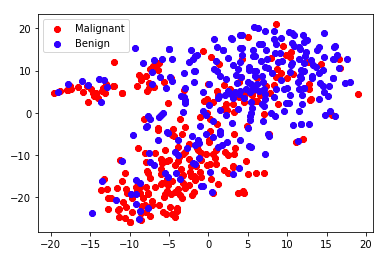}
  %\caption{5a}
  %\label{fig:sfig3}
%\end{subfigure}%
%\begin{subfigure}{.5\textwidth}
  %\centering
  \includegraphics[width=.95\columnwidth]{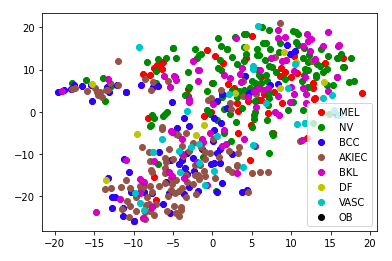}
  %\caption{5b}
  %\label{fig:sfig4}
%\end{subfigure}
\caption{$t$-SNE Map. Each point in the figure represents a test-set lesion separately colored by malignancy (Top) and lesion sub-type (Bottom): Melanoma (MEL), Melanocytic Nevus (NV), Basal Cell Carcinoma (BCC), Actinic Keratosis/Bowen's Disease (AKIEC), Benign Keratosis (BKL), Dermatofibroma (DF), Vasular Lesion (VASC) and Other Benign (OB).}
\label{fig: t-SNE for test dataset}
\end{figure}

\subsubsection{Dermoscopy Data Augmentation}
Finally, we consider whether augmenting the discovery dataset with the publicly available ISIC2018 dataset improves the performance characteristics of the proposed model.
Specifically, the ISIC20128 (training) dataset which consists of only dermoscopy images is meant to compensate for the low representation of dermoscopy images in our discovery dataset, \emph{i.e.}, only 11\% of the discovery images are dermoscopy.
Results in Table~\ref{tab:ISIC} are stratified by image type (all images, smartphone (wide-field) only and dermoscopy only) are presented for three different models: $i$) malignancy prediction (assuming the positions of the lesions are available); $ii$) direct image-level classification; and $iii$) the two-stage approach with \emph{one-class} lesion identification.
As expected, data augmentation consistently improve the performance metrics of all models considered.

% The ISIC dataset we used is a seven class classification dataset. We firstly convert the 7 classes into 2 classes as we did before: if the label of a ISIC image is belonged to classes Melanoma (0), Basal cell carcinoma (2), Actinic keratosis / Bowen's disease (intraepithelial carcinoma) (3), we changed its label as malignant, otherwise we changed its label as benign. Then we mix it with our training datasets as new training datasets for both malignancy prediction model and full image classification model. The table \ref{tab:ISIC effects} shows the performance comparison between models trained on original training dataset and models trained on mix training dataset for different tasks. The results indicate adding ISIC dataset can improve the performance of models on our dataset.

\subsection{Qualitative Results}
Figure~\ref{fig:detection examples} shows examples of the \emph{one-class} lesion identification model described in Section~\ref{sc:identification}.
Note that the model is able to accurately identify lesions in images with vastly different image sizes, for which the lesion-to-image ratio varies substantially.
We attribute the model ability to do so to the FPN network that allows to obtain image representations (features) at different resolution scales.
Further, in Figure~\ref{fig: t-SNE for test dataset} we show through a two-dimensional $t$-SNE map \cite{van2008visualizing} that the representations produced by the lesion detection model (combined backbone and FPN features) roughly discriminate between malignant and benign lesions, while also clustering in terms of lesion types.

\section{Discussion}\label{sc:discussion}
The early skin lesion classification literature used largely high-quality clinical and dermoscopy images for proof of concept.
However, usability of these algorithms in the real-world remains questionable and must be tested prospectively in clinical settings.
Consumer-grade devices produce images of variable quality, however, this approach mimics the clinical work flow and provides a universally applicable image capture for any care setting.
The utility of wide-field clinical images taken with smartphone was recently demonstrated by Soenksen \emph{et. al} for detection of ``ugly duckling'' suspicious pigmented lesions \emph{vs.} non-suspicious lesions with 90.3\% sensitivity (95\% CI: 90.0-90.6) and 89.9\% specificity (95\% CI: 89.6 - 90.2) validated against three board certified dermatologists \cite{soenksen2021using}.
This use case demonstrates how clinical work flow in dermatology can be replicated with ML-based CDS.
However, the limitation is that the number needed to treat (NNT) for true melanoma detection from pigmented lesion biopsies by dermatologists is 9.60 (95\% CI: 6.97-13.41) by meta-analysis \cite{petty2020meta}. 
Hence, the task of detecting suspicious pigmented lesions should be compared against histological ground truth rather than concordance with dermatologists, for improved accuracy and comparability of model performance.
Furthermore, pigmented lesions are a small subset of the overall task to detect skin cancer, as melanomas constitute fewer than 5\% of all skin cancers.
Our approach utilizing wide-field images to detect lesions of interest demonstrated encouraging mAP, IoU and Recall metrics, considering the sample size used.
This primary step is critical in the clinical workflow where images are captured for lesions of interest but lesion annotation is not possible in real time.
An ideal ML-based CDS would identify lesion of interest and also provide the likelihood of malignancy and the sub-type annotations as feedback to the user.
Our study demonstrates malignancy classification for the three most common skin cancers (BCC, SCC and Melanoma) \emph{vs.} benign tumors with smartphone images (clinical and dermoscopy) with encouraging accuracy when validated against histopathological ground truth.
The usability of this algorithm is further validated by comparison with dermatologists with variable levels of dermoscopy experience, showing comparable performance to dermatologists in both clinical and dermoscopy binary classification tasks, despite low dermoscopy image data (11\%) in the Discovery set.
This two-stage model, with the current performance level, could be satisfactorily utilized in a PCP triage to dermatology (pending prospective validation) at scale for images concerning for malignancy as a complete end-to-end system.
Interestingly, the additional ISIC high-quality dataset (predominantly dermoscopy images) improved performance across both clinical and dermoscopy image sets.
This suggests that smartphone image data can be enriched by adding higher quality images.
It is unclear if this benefit is due to improvement in image quality or volume, and remains an area of further study. 

Finally, we demonstrated that comprehensive demographic and clinical data is not critical for improving model performance in a subset of patients, as the image classification model alone performs at par with the combination model.
Clinicians often make contextual diagnostic and management decisions when evaluating skin lesions to improve their accuracy.
Interestingly, this clinical-context effect that improves diagnostic accuracy at least in pigmented lesions maybe dependent on years of dermoscopy experience \cite{haenssle2018man}.
The value of clinical context in model performance has not been studied extensively and remains an area of further study in larger datasets.

\paragraph{Limitations}
Limitations of the study include a small discovery image dataset, predominantly including light and medium skin tones, and with less than 2\% of images included with dark skin tone.
However, this may represent the bias in the task itself as skin cancers are more prevalent in light- followed by medium-skin tones.
Given the large range of skin types and lesions encountered in clinical practice, additional images may improve performance and generalizability.
At scale, image data pipelines with associated metadata are a key resource needed to obtain inclusive ML-based CDS for dermatology. 
Improved image quality and/or volume improves performance as demonstrated by the ISIC dataset incorporation into the model, however, this theoretical improvement in performance needs validation in prospective clinical settings. 
While the pure clinical model incorporates a comprehensive list and accounts for temporal association of this metadata with detection of lesions, it is not an exhaustive list as it does not include social determinants such as sun-exposure behavior and tanning bed usage; two critical factors contributing to increasing incidence of skin cancer.
In particular, metadata including lesion symptoms and evolution is missing and should be incorporated in future studies.
Finally, it should be noted that lesions included in this study were evaluated and selected for biopsies in dermatology clinics.
If this model was to be utilized in other clinical settings such as primary care, additional validation will be needed as pre-test probability of lesion detection is different among clinical settings \cite{petty2020meta}.

\section*{Acknowledgements}
The authors would like to thank Melodi J. Whitley, Ph.D., MD (MJ) and Amanda Suggs, MD (AS) for their assistance with the manual classification of images for the test dataset; and the Duke Institute for Health Innovation (DIHI) for providing access to the clinical data.

\bibliographystyle{unsrt}
\biboptions{authoryear}
\bibliography{main_arxiv}

\end{document}